\documentclass[runningheads]{llncs}
\usepackage{hyperref}
\usepackage{graphicx}
\usepackage[utf8]{inputenc}
\usepackage{newunicodechar}
\newunicodechar{²}{\ensuremath{{}^2}}

\begin{document}
\title{ISLES 2022: A multi-center magnetic resonance imaging stroke lesion segmentation dataset}

\titlerunning{ISLES 2022: A multi-center MRI stroke lesion segmentation dataset}
%
\author{Moritz Roman Hernandez Petzsche \inst{1} \and
Ezequiel de la Rosa \inst{2,3} \and
Uta Hanning\inst{4} \and
Roland Wiest\inst{5} \and
Waldo Enrique Valenzuela Pinilla \inst{5} \and
Mauricio Reyes \inst{6} \and
Maria Ines Meyer \inst{2} \and
Sook-Lei Liew \inst{7} \and
Florian Kofler \inst{1,3,8} \and
Ivan Ezhov \inst{1,3,8} \and
David Robben \inst{3,9,10} \and
Alexander Hutton \inst{7} \and
Tassilo Friedrich \inst{1} \and
Teresa Zarth \inst{1} \and
Johannes Bürkle \inst{1} \and
The Anh Baran \inst{1} \and
Bjoern Menze \inst{3,11} \and
Gabriel Broocks \inst{4} \and
Lukas Meyer \inst{4} \and 
Claus Zimmer \inst{1} \and
Tobias Boeckh-Behrens \inst{1} \and
Maria Berndt \inst{1} \and
Benno Ikenberg \inst{12} \and
Benedikt Wiestler \inst{1,8} \and
Jan S. Kirschke \inst{1,8}}

\authorrunning{Hernandez Petzsche et al.}

\institute{Department of Diagnostic and Interventional Neuroradiology, Klinikum rechts der Isar, School of Medicine, Technical University of Munich, Germany \and
icometrix, Leuven, Belgium \and Department of Informatics, Technical University of Munich, Munich, Germany \and Department of Diagnostic and Interventional Neuroradiology, University Medical Center Hamburg-Eppendorf, Germany \and Institute of Diagnostic and Interventional Neuroradiology, University of Bern, Bern, Switzerland \and ARTORG Center for Biomedical Engineering Research, Univ. of Bern, Switzerland \and Chan Division of Occupational Science and Occupational Therapy, University of Southern California, Los Angeles, CA, USA \and TranslaTUM – Central Institute for Translational Cancer Research, Technical University of Munich, Munich, Germany \and Medical Imaging Research Center (MIRC), KU Leuven, Leuven, Belgium \and Medical Image Computing (MIC), ESAT-PSI, Department of Electrical Engineering, KU Leuven, Leuven, Belgium \and Department of Quantitative Biomedicine, University of Zurich, Zurich, Switzerland \and Department of Neurology, Klinikum rechts der Isar, School of Medicine, Technical University of Munich, Germany}
\maketitle              
\begin{abstract}
Magnetic resonance imaging (MRI) is a central modality for stroke imaging. It is used upon patient admission to make treatment decisions such as selecting patients for intravenous thrombolysis or endovascular therapy. MRI is later used in the duration of hospital stay to predict outcome by visualizing infarct core size and location. Furthermore, it may be used to characterize stroke etiology, e.g. differentiation between (cardio)-embolic and non-embolic stroke. Computer based automated medical image processing is increasingly finding its way into clinical routine. Previous iterations of the Ischemic Stroke Lesion Segmentation (ISLES) challenge have aided in the generation of identifying benchmark methods for acute and sub-acute ischemic stroke lesion segmentation. Here we introduce an expert-annotated, multicenter MRI dataset for segmentation of acute to subacute stroke lesions. This dataset comprises 400 multi-vendor MRI cases with high variability in stroke lesion size, quantity and location. It is split into a training dataset of $n=250$ and a test dataset of $n=150$. All training data will be made publicly available. The test dataset will be used for model validation only and will not be released to the public. This dataset serves as the foundation of the ISLES 2022 challenge with the goal of finding algorithmic methods to enable the development and benchmarking of robust and accurate segmentation algorithms for ischemic stroke.

\end{abstract}
\section{Background \& Summary}

Stroke is a leading cause of morbidity and mortality worldwide \cite{lozano2012global}. Up to two thirds of stroke survivors suffer permanent disability \cite{feigin2014global}. In the last decade, the advent of endovascular reperfusion therapy has significantly improved stroke outcome in patients with large vessel occlusions \cite{berkhemer2015randomized,goyal2015randomized,jovin2015thrombectomy,saver2015stent}. Image-based guidance of revascularization treatment decisions has further improved patient outcome for computer tomography (CT) \cite{albers2018thrombectomy,campbell2015endovascular,ma2019thrombolysis} and magnetic resonance imaging (MRI) \cite{hjort2005magnetic,thomalla2011dwi}. Computer aided image analysis, especially for CT perfusion data has already found entry into clinical routine in many centers and is recommended by national guidelines \cite{ringleb2021akuttherapie} to aid decision making regarding reperfusion therapy \cite{rava2020assessment,xiong2019comparison,mokin2017predictive,rava2021assessment}. Deep learning approaches have been shown to facilitate clinical interpretation of CT perfusion data \cite{clerigues2019acute,hakim2021predicting,robben2020prediction,de2020differentiable,de2021aifnet}. Segmentation based volumetric analyses of stroke lesions in magnetic resonance imaging (MRI) are often performed for research purposes and have been shown to predict ischemia outcome \cite{meng2021infarct,zecavati2014utility,freyschlag2019volume}. However, stroke lesion segmentations are usually painstakingly performed by hand and the quality of annotations are heavily dependent on preexisting neuroimaging experience of the rater and the total time and effort invested. The time-consuming nature of this task prevents the regular use of segmentations during clinical routine, which is further impeded by the high inter-observer variability. Automated annotation of stroke lesions could be used in clinical routine to guide therapeutic decisions in an acute setting and to predict outcome at the subacute to chronic stage. Stroke lesion segmentation could also be used to automatically classify stroke etiology in post-stroke MRI.

The first Ischemic Stroke Lesion Segmentation (ISLES) challenge, which took place in 2015, was split into two sub-challenges: Sub-acute Stroke Lesion Segmentation (SISS) and Stroke Perfusion Estimation. The goal of SISS (with a total of 64 cases for training and testing) was to segment subacute stroke lesions using conventional post-stroke MRI sequences, including T2 and T1 weighted imaging, fluid attenuated inversion recover (FLAIR), and diffusion weighted imaging (DWI) \cite{maier2017isles}. The ISLES challenge 2018, its second iteration, was set up to predict infarct core in DWI using CT perfusion data \cite{hakim2021predicting}. Both ISLES events received major attention from the research community: there were 120 database downloads until the ISLES15 challenge day with 14 participating teams, and the number of participating teams was roughly duplicated in the latest ISLES'18 edition. The ISLES'15 and ISLES'18 challenges played a crucial role in identifying prominent methods for acute and sub-acute ischemic stroke lesion segmentation. These datasets have since served as important benchmarks for the scientific community.

Based on the experience gained from these previous editions, ISLES'22 aims to benchmark acute and sub-acute ischemic stroke MRI segmentation using 400 cases. This MICCAI 2022 challenge edition is organized with the task of DWI infarct segmentation in acute and sub-acute stroke. ISLES'22 differs in several ways from the previous challenge editions in ischemic stroke by: 1) targeting the delineation of not only large infarct lesions, but also of multiple embolic and/or cortical infarcts (typically seen after mechanical recanalization), and 2) by evaluating both pre- and post- interventional MRI images. Altogether, ISLES'22 provides a considerably larger dataset than before (more than 5x the caseload than in the ISLES'15 edition). For detailed information about the ISLES'22 challenge event, readers are referred to \cite{ezequiel_de_la_rosa_2022_6362388}. This challenge is held together with the ATLAS challenge (\href{https://atlas.grand-challenge.org/}{https://atlas.grand-challenge.org/} \cite{liew2021large} using the web-based platforms \href{http://www.isles-challenge.org/}{http://www.isles-challenge.org/} and \href{https://isles22.grand-challenge.org/}{https://isles22.grand-challenge.org/}. 

From a clinical perspective, ISLES'22 focuses on the clinical growing interest of acute embolic infarct patterns, both pre-intervention (i.e. at a very early disease state) and post-intervention (typical, post-interventional sub-acute infarct patterns not restricted to a single vessel territory). This clinical problem also challenges the participants from a technical perspective: teams will deal with a wider ischemic stroke disease spectrum, involving variable lesion size and burden, more complex infarct patterns and variable anatomically located lesions in data from multiple centers. The diversity of the ISLES'22 dataset will provide a unique challenge for participants.

\section{Methods}

\subsection{Ethical Statement}
This retrospective evaluation of imaging data was approved by the local ethics boards of all participating centers. 

\subsection{Subject selection}

Inclusion criteria for the dataset: Subjects 18 years or older who had received MR imaging of the brain for previously diagnosed or suspected stroke were included in this study. The imaging protocol required at least a FLAIR and DWI sequence. DWI consists of a trace image at a b-value up to 1000 s/mm² as a well as its corresponding apparent diffusion coefficient (ADC) map. Image acquisition was performed on one of the following devices: 3T Philips MRI scanners (Achieva, Ingenia), 3T Siemens MRI scanner (Verio) or 1.5T Siemens MAGNETOM MRI scanners (Avanto, Aera). MRIs included were intentionally chosen to be heterogeneous to ensure the best possible training and generalization of the algorithms. See Table \ref{Table_MRI} for an overview of MRI acquisition parameters. Images were obtained by healthcare professionals as part of the clinical imaging routine for stroke patients at three different stroke centers.

\begin{table}
    \centering

\caption{Overview of MRI imaging parameters. Range of values [Min-Max] are shown. TR: Repetition time; TE: Echo time; TI: Inversion time.}\label{Table_MRI}
\begin{tabular}{cccccccc}

{\large Modality}               & {\large FLAIR}               & {\large DWI}                 \\
\hline
TR   (ms)              & {[}4800-12000{]}    & {[}3175-16439{]}    \\
TE   (ms)              & {[}103-395{]}       & {[}55-91{]}         \\
TI   (ms)              & {[}1650-2850{]}     & -                   \\
Flip   Angle ($\circ$)       & {[}90-180{]}        & 90                  \\
Voxel   (mm2)          & {[}0.23x0.23-1x1{]} & {[}0.88x0.88-2x2{]} \\
Slice   thickness (mm) & {[}0.7-9.6{]}       & {[}2.0-6.5{]}       \\
Gap   (mm)             & {[}0.7-6.5{]}       & {[}2.0-5.0{]} \\ 
\hline
\end{tabular}
\end{table}

\begin{figure}[!t]
\includegraphics[width=\textwidth]{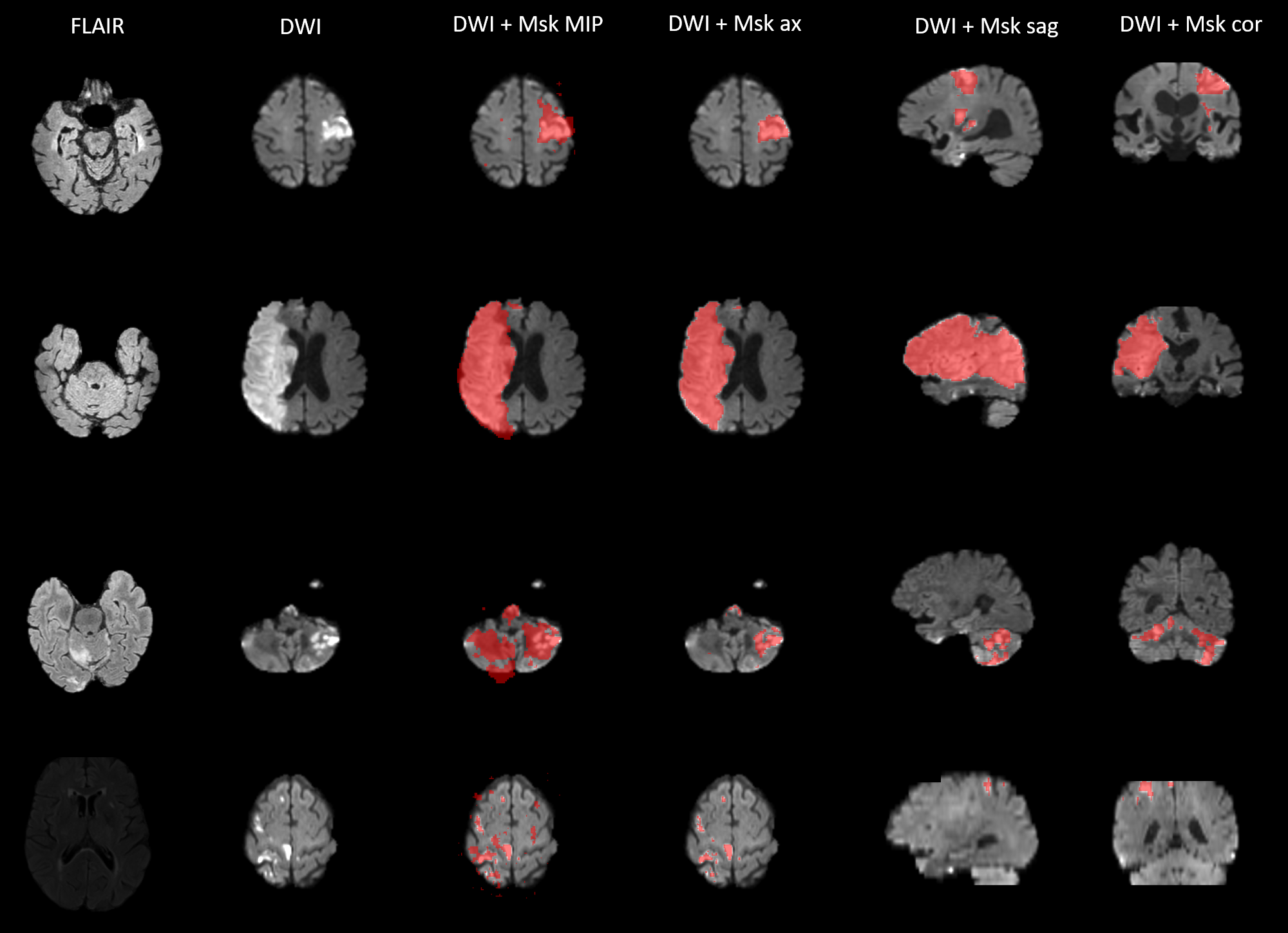}
\caption{Exemplary 3D Snapshots through the ischemic center of mass. Axial FLAIR and DWI images are displayed in the top leftmost columns. The 3rd column from the left shows the maximum intensity projection (MIP) of the mask (Msk). The three rightmost columns show DWI with mask overlay (without MIP) in the three anatomical planes.
Top row: Larger left-sided infarct with bilateral punctiform, likely embolic satellite ischemias.
2$^{nd}$ row: Infarct of the entire right sided middle cerebral artery territory.
3$^{rd}$ row: Bilateral cerebellar and occipital infarcts after posterior circulation ischemia.
Bottom row: Bilateral punctiform infarcts, likely resulting from multiple micro-embolic occlusions. } \label{fig:segm_examples}
\end{figure}

Care was taken to select a broad spectrum of infarct patterns. All vascular territories were included at a similar rate. A large subset of patients with posterior circulation ischemia (e.g. due to basilar artery occlusion) was included in this study. Due to a high degree of closely situated scull-base artefacts in DWI, infratentorial ischemias are difficult to segment for unexperienced raters and are frequently overlooked by segmentation algorithms with little training exposure to posterior circulation ischemia. These additional labeling challenges lead us to include more cases of posterior circulation ischemia than would be expected if case selection were truly random. Figure \ref{fig:segm_examples} shows sample cases portraying the spectrum of ischemia included in this dataset.

Segmentation difficulties also frequently arise in cases with large amounts of punctiform infarcts. For human raters, to adequately capture the entire ischemic territory, a time-consuming segmentation effort is required. Similarly, machine trained algorithms frequently fail to capture all the small affected regions. In $n=5$ patients, where MR imaging was acquired for suspected stroke, no ischemia was found. We chose to include this small subset to diversify the dataset. Table \ref{Tab_lesions} gives an overview of infarct volumes per scan and per lesion, as well as the number of unconnected ischemias per scan in the entire dataset. The number of unconnected ischemias has been calculated using the python library cc3d \cite{silversmith_william_2021_5719536}.

\begin{table}
\centering
\caption{Overview of MRI imaging parameters. Range of values are shown. TR: Repetition time; TE: Echo time; TI: Inversion time.}\label{Table_lesions}
\begin{tabular}{cccc}
&  \shortstack{Scan infarct \\ volume (ml)} & \shortstack{Number of  \\ unconnected ischemias} & \shortstack{Lesion-wise \\ infarct volume (ml)} \\
\hline
Mean (std) & 22.532 (44.756) & 9.298 (13.496) & 2.423 (15.804) \\
{[}Min, Max{]} & {[}0.000, 482.152 {]}& {[}0, 126{]} & {[}0.003, 477.264 {]}\\        
\hline
\end{tabular}
\label{Tab_lesions}
\end{table}

In the hyper-acute phase of ischemic stroke, up to 4.5 hours post onset, restricted diffusion is present (high signal on DWI and low signal on ADC) while the FLAIR in the affected parenchyma remains without changes. This imaging phenomenon is called a FLAIR-DWI mismatch and is used in clinical practice to estimate the time window in patients where the time of onset is unknown. An accurate estimate of the time of onset is crucial to make decisions regarding revascularization treatment \cite{thomalla2011dwi}.  In acute ischemic stroke, following the hyper-acute phase and usually defined in literature as 0 to 7 days from onset, DWI and FLAIR show a high signal with reduced ADC values in the affected brain parenchyma. In the subacute stage, between 1 to 3 weeks post onset, high DWI signal begins to diminish while ADC first normalizes to values of healthy brain tissue, a phenomenon frequently referred to as pseudonormalization. FLAIR signal remains high. In the chronic stage, beginning 3 weeks after onset, DWI signal is variable but usually iso- to hypo-intense depending on underlying T2 signal, while ADC values are high \cite{allen2012sequence,lansberg2001evolution,warach1992fast,warach1995acute,lutsep1997clinical,schlaug1997time,nagesh1998time,schwamm1998time,yang1999serial,beaulieu1999longitudinal}. MRIs with late acquisition post stroke ($>$ 1 week) often lead to a decreased DWI signal intensity for ischemic brain parenchyma. A lower signal intensity in DWI leads to lower MRI sensitivity for stroke and segmentation difficulty for both human and machine raters. In these cases, it is especially difficult to adequately annotate the border between ischemia and healthy brain tissue. This dataset includes cases of MRIs in various stage of sub-acute stroke from multiple previous studies to find machine learning solutions to this frequent issue in stroke lesion segmentation \cite{schonfeld2020effect,schonfeld2020sub}. 

\subsection{Ground truth stroke lesion segmentation}

\begin{figure}[!b]
\includegraphics[width=\textwidth]{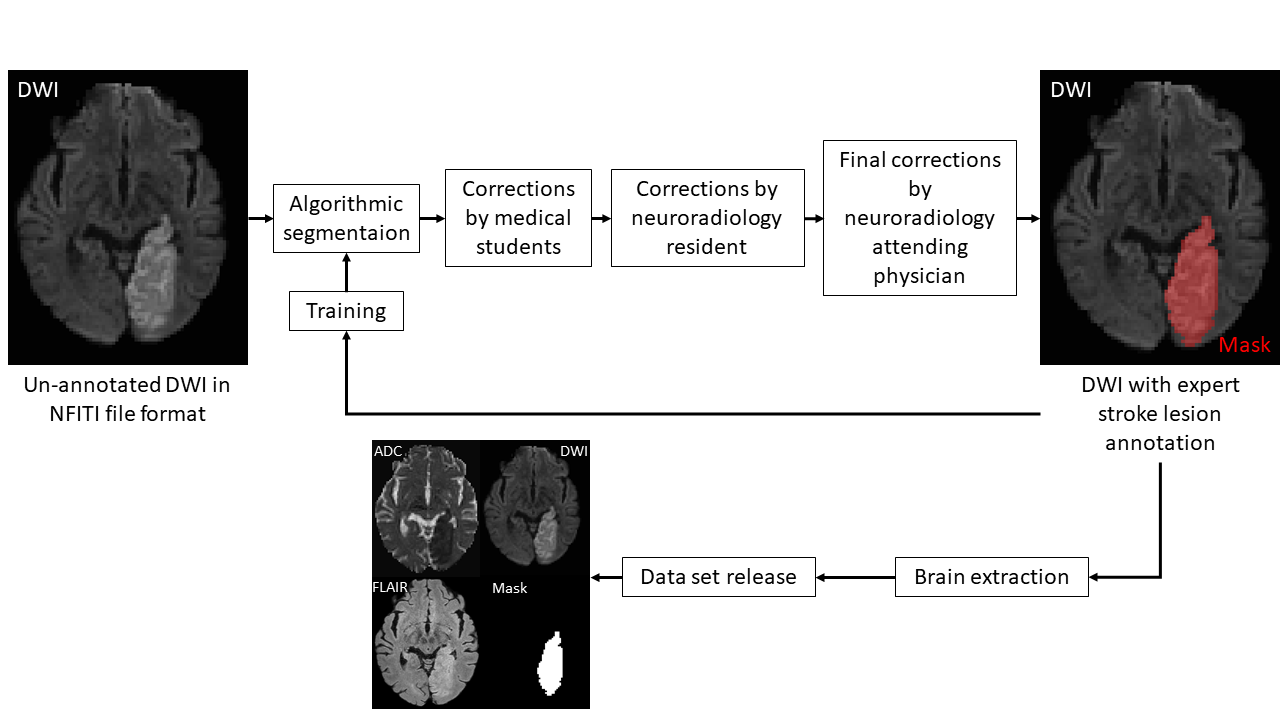}
\caption{Workflow of the hybrid human-algorithm stroke lesion segmentation applied in this dataset.
} \label{fig:Workflow}
\end{figure}

A hybrid human-algorithm annotation scheme was applied.  First, the MR input data was anonymized by conversion to Neuroimaging Informatics Technology Initiative (NIfTI) format (\href{https://nifti.nimh.nih.gov/nifti-1}{https://nifti.nimh.nih.gov/nifti-1}), according to the Brain Imaging Data Structure (BIDS) convention (\href{https://bids.neuroimaging.io/}{https://bids.neuroimaging.io/}). As part of pre-processing, DWI and its corresponding ADC map were resliced using ANTs \cite{avants2011reproducible} to an axial isotropic voxel size of 2$\times$2 $mm^2$. This was performed only for the cases of Center \#1 (see below), as this represents the original acquisition voxel size. The slice thickness remained as acquired. FLAIR data remained as exported. A 3D UNet \cite{cciccek20163d} was trained on DWI data that was pre-annotated for other research projects by using a single MRI modality (B=1000 DWI). This algorithm was trained in-house with labeled data from University Hospital of Munich. Later, this algorithm pre-segmented all scans intended for later release. All algorithmic segmentations were then checked and edited by medical students with special stroke lesion segmentation training. Their annotations were in turn revised by a neuroradiology resident. Due to inadequate performance of the primary pre-segmentation algorithm, largely due to the reasons mentioned above, the original training data was critically revised and a large part was deemed inadequate and discarded. A new pre-segmentation algorithm was trained on annotations deemed correct or edited by the neuroradiologist in training. This resulted in more accurate predictions and a lesser effort of correcting annotations by medical students. All medical-student edited annotations were critically revised by the neuroradiologist in training and a large part of the final data set were reviewed by one out of three attending neuroradiologists, all of them with more than 10 years of experience in stroke imaging. Manual stroke lesion segmentations were performed using ITK-Snap or 3D Slicer open-source tools for brain imaging visualization and segmentation \cite{yushkevich2006user,fedorov20123d}. Figure \ref{fig:Workflow} shows an overview of the annotation workflow followed in preparation of the release of the dataset.

\subsection{Post-annotation Data Pre-processing}

In preparation for data-set release and in accordance to the ethical approval obtained for this challenge, the imaging data was irreversibly anonymized. For this de-identification of patients, brain-extraction was performed mask-based using the HD-BET algorithm \cite{isensee2019automated} completion of the annotation process. FLAIR to DWI rigid registration was performed using Elastix \cite{klein2009elastix,shamonin2014fast} and subsequent skull-stripping of DWI and ADC using the registered brain mask was performed. After skull stripping for de-identification, all imaging sequences were returned to their native space before data release.

\subsection{Data Records}

Data structure and file formats. All medical imaging files were exported from the Picture Archiving and Communication System in the NIfTI format. Segmentation masks are also saved in NIfTI format. All data in the ISLES’22 dataset was separated into a training dataset (250 subjects) and a test dataset (150 subjects). Corresponding scanner metadata from the Digital Imaging and Communications in Medicine (DICOM) header in the JSON file format is provided with the datasets if available. MRIs from the following centers were included:

\begin{itemize}
    \item [$\ast$] Center \#1: University Hospital of the Technical University Munich, Munich, Germany.
    
    \item [$\ast$] Center \#2: University Hospital of Bern, Bern, Switzerland.
    
    \item [$\ast$] Center \#3: University Medical Center Hamburg-Eppendorf, Hamburg, Germany.
\end{itemize}

The train set comprises data from centers \#1 and \#2. The test set comprises data from all the three centers in equal parts as follows:
Acute to early sub-acute stroke data from centers \#1 and \#3 (MRIs acquired after revascularization therapy).
Hyper-acute to acute stroke data from center \#2 (MRIs acquired before revascularization therapy).
Thus, in this ISLES’22 task we will evaluate the robustness and generalization capability of the proposed models over 1) new scans coming from two centers already used at training stage, 2) new scans coming from a new (unseen at training stage) center, and 3) new scans acquired before revascularization therapy, coming from a center already seen at training stage. The split between training and test data set will be done so that both sets will include a similar variance of stroke lesion patterns ranging from large territorial infarcts to small punctiform ischemia.  

\subsection{Technical Validation} 

The presented medical imaging data was derived from the picture archiving and communication system of the corresponding institution and therefore fully complies with the legal standards and quality controls for the acquisition of medical imaging in Germany, European Union and Switzerland, as well as the industrial standards of the scanner vendors. Segmentation masks were prepared and annotated at voxel-level by a human-machine hybrid algorithm with hierarchical manual checks and corrections first by specifically trained medical students and later a neuroradiology resident. Afterwards, the masks were reviewed, corrected, and finally approved by an expert neuroradiologist. 

\subsection{Data and Code Availability}

The ISLES 2022 data can be downloaded by following the steps described at  \href{https://isles22.grand-challenge.org/dataset/}{https://isles22.grand-challenge.org/dataset/}. This data is published under the creative commons license CC BY-SA 4.0. Upon reasonable request, respecting patient confidentiality and ethical standards, special arrangements for use and re-distribution can be made directly with the authors. 
In order to facilitate future users of this dataset to get familiarized with the images, we have released the following ISLES 2022 Github repository: \href{https://github.com/ezequieldlrosa/isles22}{https://github.com/ezequieldlrosa/isles22}. The repository contains scripts to read the images, visualize them, and to quantify the algorithmic results performance with the same metrics used in the challenge to rank participants.


\bibliography{mybibfile}

\end{document}